\pdfoutput=1
\documentclass{article}

\usepackage{times}
\usepackage{graphicx} % more modern
\usepackage{subfigure} 

\usepackage{natbib}

\usepackage{algorithm}
\usepackage{algorithmic}

\usepackage{url}
\usepackage{amsmath,amssymb}
\usepackage{amsfonts}
\usepackage{amsthm,mathrsfs}
\usepackage{widetext}
\providecommand{\norm}[1]{\lVert#1\rVert_{\mathcal{F}}}
\DeclareMathOperator{\trans}{^{\mathrm{T}}}

\providecommand{\scalProd}[2]{\langle#1#2\rangle_{\mathcal{F}}}

\DeclareMathOperator{\tr}{Tr}

\newtheorem{theorem}{Theorem}

\begin{document} 

%\twocolumn[
%\icmltitle{Supervised LogEuclidean Metric Learning\\
%  for Symmetric Positive Definite Matrices}
%
%% It is OKAY to include author information, even for blind
%% submissions: the style file will automatically remove it for you
%% unless you've provided the [accepted] option to the icml2015
%% package.
%\icmlauthor{Florian Yger}{florian@ms.k.u-tokyo.ac.jp}
%\icmladdress{Department of Complexity Science and Engineering, Graduate School of Frontier Sciences, \\The University of Tokyo,\\ 7-3-1 Hongo, Bunkyo-ku, Tokyo 113-0033, Japan}
%\icmlauthor{Masashi Sugiyama}{sugi@k.u-tokyo.ac.jp}
%\icmladdress{Department of Complexity Science and Engineering, Graduate School of Frontier Sciences,\\ The University of Tokyo,\\ 7-3-1 Hongo, Bunkyo-ku, Tokyo 113-0033, Japan}
%% You may provide any keywords that you 
%% find helpful for describing your paper; these are used to populate 
%% the "keywords" metadata in the PDF but will not be shown in the document
%%\icmlkeywords{boring formatting information, machine learning, ICML}
%\icmlkeywords{metric learning, Symmetric Positive Definite Matrices, LogEuclidean distance}
%
%\vskip 0.3in
%]

\title{Supervised LogEuclidean Metric Learning \\ for Symmetric Positive Definite Matrices}
\author{Florian Yger \and Masashi Sugiyama\\  Department of Complexity Science and Engineering\\ Graduate School of Frontier Sciences\\ The University of Tokyo}
\date{February 6th, 2015}
\maketitle

\begin{abstract} 
Metric learning has been shown to be
highly effective to improve the performance of nearest neighbor classification.
In this paper, we address the problem of metric learning 
for \emph{symmetric positive definite} (SPD) matrices such as covariance matrices,
which arise in many real-world applications.
Naively using standard Mahalanobis metric learning methods under the Euclidean geometry
for SPD matrices is not appropriate,
because the difference of SPD matrices can be a non-SPD matrix
and thus the obtained solution can be uninterpretable.
To cope with this problem, we propose to use a properly parameterized \emph{LogEuclidean distance}
and optimize the metric with respect to \emph{kernel-target alignment},
which is a supervised criterion for kernel learning.
Then the resulting non-trivial optimization problem is solved by utilizing
the \emph{Riemannian geometry}.
Finally, we experimentally demonstrate the usefulness of our LogEuclidean metric learning algorithm 
on real-world classification tasks for EEG signals and texture patches.
\end{abstract}

\section{Introduction}
Defining a distance to compare data points is an important problem occurring in many machine learning tasks. For instance, in classification, the nearest neighbor method~\cite{cover1967nearest} is equipped with a distance to identify the nearest neighbors. The performance of such a classifier highly depends on the quality of the equipped distance,
and hence automatically finding a relevant distance from data has been the aim of many metric learning algorithms. 
So far, various methods have been developed
for learning \emph{Mahalanobis distances} in the Euclidean setting,
and such metric learning methods have been successfully applied to 
diverse real-world problems including music recommendation~\cite{
%mcfee2012learning, 
lim2013robust}, face recognition~\cite{lu2012neighborhood},
image classification~\cite{mensink2012metric}, 
link prediction in networks~\cite{shaw2011learning} and bioinformatics~\cite{wang2012prodis}. 
For the survey of recent advances, issues and perspectives in metric learning,
see \citet{bellet2013survey}. 

Classical metric learning approaches focused on an Euclidean setting
which do not apply to complex or structured objects.
Recently, metric learning for complex objects such as
histograms~\cite{cuturi2014ground, kedem2012non}, %wang2012supervised}, 
binary codes~\cite{norouzi2012hamming} and
strings~\cite{bellet2011learning} have been actively explored,
to which the Euclidean distance is not relevant.
However, to the best of our knowledge,
metric learning for
\emph{symmetric positive definite} (SPD) matrices
has not been investigated thoroughly.
Learning from SPD matrices,
particularly from \emph{covariance matrices},
arises in a number of important classification tasks
such as brain imaging~\cite{arsigny2006log,dryden2009non},
brain-computer interfaces~\cite{barachant2010riemannian,barachant2013classification},
pedestrian detection~\cite{tuzel2008pedestrian}
and texture classification~\cite{tuzel2006region,tou2009gabor}. 
In these applications, covariance matrices are either extracted from
a physical model of the studied phenomenon
(for diffusion tensor imaging)
or as an empirical estimator from observations
(for signal processing and computer-vision tasks).
The purpose of this paper is to investigate
metric learning for SPD matrices
to boost the classification performance.

\begin{figure}[t]
\centering
    \includegraphics[width=0.65\linewidth]{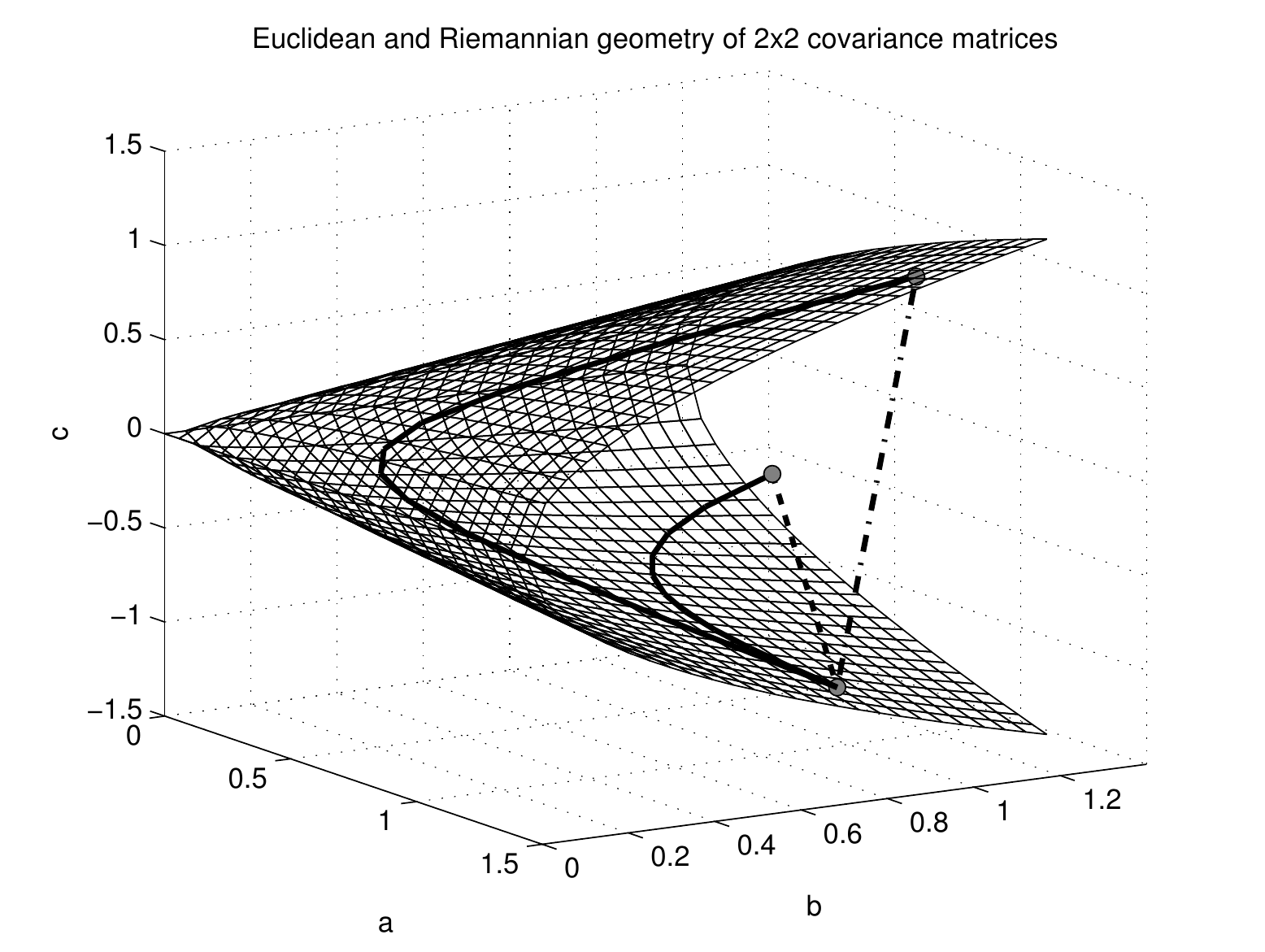}
\caption{Comparison between Euclidean (straight dashed lines) and Riemannian (curved solid lines) distances measured between points of the space $\mathcal{P}_2$.}
\label{fig:psd2d}
\end{figure}

SPD matrices belong to an Euclidean\footnote{Note that, in order to be an Euclidean space, the space should be equipped with the Frobenius inner product $\langle A, B \rangle_{\mathcal{F}} = \tr(A^\top B)$ and the derived norm $||A||_{\mathcal{F}} = \sqrt{\langle A, A \rangle_{\mathcal{F}}}$.}
space.
%   (namely the space of symmetric matrices).
For example, $2\times2$ SPD matrix $A$ can be written as ${A = \left[ \begin{array}{cc}
a & b \\
b & c  \end{array} \right]}$ with ${ac - b^2 > 0 }$, $a>0$ and $c>0$.
Then symmetric matrices can be represented as points in $\mathbb{R}^3$ and
the constraints can be plotted as a cone, inside which SPD matrices lie strictly
(see Fig.~\ref{fig:psd2d}). 
A straightforward approach for learning a metric in this space would be 
to simply use the Euclidean distance $\delta_e$:
\begin{align}
 \delta_e (A,B)& = \norm{ A -B}\label{eq:euclDist},
\end{align}
where $\norm{\cdot}$ denotes the Frobenius norm.
The Euclidean geometry of symmetric matrices implies
that distances are computed along straight lines 
according to $\delta_e$
(see Fig.~\ref{fig:psd2d} again).
%  (as defined in Eq.~\ref{eq:euclDist}).  

% Hence the set of $d \times d$ positive definite matrices is denoted by $\mathcal{P}_d$

% The difference between Euclidean and Riemannian geometry (discussed in Section~\ref{sec:discussion}) for the space $\mathcal{P}_d$ is illustrated in Fig.~\ref{fig:psd2d}. As already shown in~\cite{fletcher2004principal}, the space $\mathcal{P}_2$ can be represented in three dimensions. In $\mathcal{P}_2$, 

However, the Euclidean geometry for averaging SPD matrices
can result in a \emph{swelling effect}~\cite{arsigny2007geometric},
i.e., the determinant of the average can be bigger than the determinant of each matrix.
As also remarked in~\citet{fletcher2004principal} and illustrated in Fig.~\ref{fig:psd2d},
this geometry creates a \emph{non-complete space},
meaning that interpolation of SPD matrices is possible,
but extrapolation can produce indefinite matrices.
Hence, the swelling effect and the non-completeness of the space
can result in uninterpretable solutions.

To avoid this problem,
we use a more natural metric to compare SPD matrices,
namely, the \emph{LogEuclidean distance} $\delta_l$:
\begin{align}
\delta_{l} (A,B) = \norm{\log \left( A\right) - \log \left( B \right)},
\label{eq:LogEDist}
\end{align} 
where $\log(\cdot)$ stands for the matrix logarithm.
The LogEuclidean metric (and the derived distance and kernel) has been used
in the literature~\cite{arsigny2006log,barachant2013classification},
but its parameterization remained underestimated until recently~\cite{yger2013review}.
In this paper, we propose a supervised approach to learning the LogEuclidean metric.
More specifically, we formulate our LogEuclidean metric learning problem
as the \emph{kernel-target alignment} problem~\cite{cristianini2001kernel,cortes2012algorithms},
and solve the non-trivial optimization problem using the Riemannian geometry for SPD matrices.
Through experiments on signal processing and computer vision applications,
we demonstrate the efficacy of our proposed metric learning method.

% The remainder of this paper is organized as follows:
% first, in Sec.~\ref{sec:proposition}, we introduce our approach
% which formulates the LogEuclidean metric learning problem
% as a non-trivial optimization problem on a matrix manifold.
% Then, in Sec.~\ref{sec:discussion}, we discuss properties
% of our LogEuclidean metric learning problem from geometric viewpoints.
% In Sec.~\ref{sec:expeNum}, we experimentally compare our proposed method 
% to baseline approaches on a toy data set and on two real-life datasets respectively
% from signal processing and computer vision applications.
% Finally, Sec.~\ref{sec:conclusion} ends this paper with some concluding remarks 
% and perspectives on this work.

\section{Proposed approach}
\label{sec:proposition}
In this section, we describe our proposed metric learning method.
% Before delving into the presentation of the problem we are addressing, let us introduce some notations. 
% \paragraph{Notations.} 

\subsection{Formulation of metric learning under LogEuclidean distance}
Let $\mathcal{S}_d$ be the set of real symmetric matrices of size $d \times d$.
A matrix $A$ is said to be positive-definite ($A \succ 0$) 
if ${ x^\top A x > 0}$ for all non-zero $x \in \mathbb{R}^d$,
and the set of $d \times d$ SPD matrices is denoted by $\mathcal{P}_d$.

To learn a metric, we need to parameterize a distance.
In this paper, following \citet{bhatia2009positive},
we focus on the \emph{congruent transform}, i.e., for any $A,G\in\mathcal{P}_d$,
%\begin{align*}
%  \Gamma_{G^{-\frac{1}{2}}}: A\to G^{-\frac{1}{2}} A G^{-\frac{1}{2}}.
%\end{align*}
\[
\begin{cases}
\quad \quad \quad \mathcal{P}_d \mapsto \quad \mathcal{P}_d\\
\Gamma_{G^{-\frac{1}{2}}}: A\to G^{-\frac{1}{2}} A G^{-\frac{1}{2}}.
\end{cases}
 \]

Thus, the LogEuclidean distance \eqref{eq:LogEDist} is parameterized
using $G\in\mathcal{P}_d$ as
\begin{align}
\delta_{l}^G (A,B)
  &= \delta_{l} \left( \Gamma_{G^{-\frac{1}{2}}} (A),\Gamma_{G^{-\frac{1}{2}}} (B) \right)
\label{eq:LogEDistG}\\
 &=\norm{\log \left( G^{-\frac{1}{2}} A G^{-\frac{1}{2}}\right) - \log \left( G^{-\frac{1}{2}} B G^{-\frac{1}{2}} \right)}. \notag
\end{align}
Our goal is to learn the parameter matrix $G$
from a set of $n$ training samples,
\begin{align*}
  \{ (X_i,y_i) | X_i \in \mathcal{P}_d, y_i\in\{+1,-1\}\}_{i=1}^n,
\end{align*}
so that the performance of the nearest neighbor classifier on $\mathcal{P}_d$ is maximally enhanced.

% In~\cite{bhatia2009positive}, for each $M \in GL (d)$ (the group of $d\times d$ invertible matrices), the author describes the congruent transformation $\Gamma_M(X)=M^{\star}X M$. This congruent transformation is a bijection from $\mathcal{P}_d$ to itself, with the property $\Gamma_{X^{-\frac{1}{2}}}(X)=\mathbb{I}_d$.

$G$ was implicitly chosen as the identity matrix in \citet{arsigny2006log},
while
$G$ was heuristically chosen as the \emph{Riemannian mean} of training samples
in~\citet{barachant2013classification} and \citet{yger2013review}.
Although these choices sound reasonable,
we argue that they are sub-optimal when $\delta_{l}^G$ is used for 
nearest neighbor classification.
Our aim is to find the optimal $G$ in terms of classification performance.

\subsection{Metric learning with kernel-target alignment}
% As previously stated, the performance of the LogEuclidean metric is greatly impacted by the choice of the hyper-parameter $G$. %cf papier MLSP
% Until now, only heuristic choices were available. Our goal is to learn a relevant hyper-parameter using a supervised criterion. In order to do so, we investigate the use of the alignment criterion (or \textit{Kernel Target Alignment} (KTA) as referred in~\cite{cristianini2001kernel} or more specifically the use of its centered counterpart~\cite{cortes2012algorithms}.

For metric learning,
we employ the (centered) \emph{kernel target alignment} (KTA) criterion \cite{cristianini2001kernel,cortes2012algorithms}: 
\begin{align*}
  \mathcal{A}(K,K^{\star}) = \frac{\scalProd{U K U, U K^{\star}U}}{\norm{U K U} \norm{U K^{\star} U}},
\end{align*}
where $K$ is a matrix to be aligned,
$K^{\star}=yy^{\trans}$ is a target matrix, $y=(y_1,\ldots,y_n)^{\trans}\in\mathbb{R}^n$,
$U=\mathbb{I}_d - \frac{1_d 1_d^\top}{d}$ is the centering matrix,
$\mathbb{I}_d$ is the identity matrix of size $d \times d$,
and $1_d$ is the $d$-dimensional vector with all ones.
For simplicity, we suppose that $y$ is centered, i.e., $Uyy^TU\to yy^T$ below.

Let $k_G\left(X,X' \right)$ be the \emph{LogEuclidean metric},
from which the LogEuclidean distance \eqref{eq:LogEDistG} is derived:
\begin{align*}
k_G\left(X,X' \right)&= \tr \left( \log \left( G^{-\frac{1}{2}} X G^{-\frac{1}{2}}  \right)  \log\left(G^{-\frac{1}{2}} X' G^{-\frac{1}{2}}  \right)  \right),
% &= \tr \left( \log \left( \Gamma_{G^{-\frac{1}{2}}}\left( X \right) \right)  \log\left(\Gamma_{G^{-\frac{1}{2}}} \left(X' \right) \right)  \right)%\label{eq:logEuclideanMetric},
\end{align*}
where $X,X',G \in \mathcal{P}_d$.
% This metric is not invariant to the effect of the transformation and our aim is to tune its parameter $G^{-\frac{1}{2}}$ in order to improve the classification performance of the distance derived from this metric.
Let $h(G)$ be the gram matrix for $k_G$:
\begin{align*}
  h_{ij}(G)=k_G(X_i,X_j).
\end{align*}
Then our metric learning problem is formulated as
\begin{align}
\max_{G\in \mathcal{P}_d}f \left( G \right) ,
\label{eq:optiPb}
\end{align}
where
\begin{align}
f \left( G \right)=\mathcal{A}(h(G), yy\trans) = \frac{\scalProd{U h(G) U,yy\trans}}{\norm{U h(G) U}}.
\label{eq:instanceKTA}
\end{align}

A naive approach to solving the above optimization problem would be
the projected gradient method, i.e.,
iteratively performing gradient ascent over $f(G)$ and
projection of the updated solution back onto $\mathcal{P}_d$.
However, since $\mathcal{P}_d$ is an open set
(the strict interior of the cone is a set of SPD matrices, see Fig.~\ref{fig:psd2d} again),
projection does not always exist.
To cope with this problem, we introduce a more sophisticated approach
based on the \emph{Riemannian geometry} below.

\subsection{Riemanian Gradient Optimization}

Considering $\mathcal{P}_d$ as a Riemannian manifold provides us useful tools for
solving our optimization problem. 
A basic optimization algorithm on the Riemannian manifold is the \emph{geodesic gradient descent}\footnote{
For more sophisticated optimization methods, see \citet{absil2009optimization} and \citet{boumal2013manopt}. },
which optimizes the objective function
along the \emph{geodesic} computed from the Riemannian gradient
$\text{grad}_G f(G)$.
For optimization problem \eqref{eq:optiPb}, the geodesic from $G_t$ to $G_{t+1}$ is given by
\begin{align}
G_{t+1}
=& G_t^{\frac{1}{2}} \exp \left(\eta G_t^{-\frac{1}{2}} \text{grad}_{G_t} f(G_t) G_t^{-\frac{1}{2}}\right) G_t^{\frac{1}{2}},
\label{geodesic}
\end{align}
where $\eta\ge0$ is the step size
and $\exp(.)$ denotes the matrix exponential.
Below, we explain how the above formula was derived.

As stated in~\citet{bhatia2009positive}, $\mathcal{P}_d$ is an open subset of the ambient Euclidean space of $d \times d$ symmetric matrices $\mathcal{S}_{d}$. As such, it is an instance of an embedded sub-manifold of  $\mathcal{S}_{d}$ and so it is a differentiable manifold of dimension $d \left( d+1 \right) /2$. 
From this structure of differential manifolds, we can derive the notion of tangent space $T_G \mathcal{P}_d$ at each point $G \in \mathcal{P}_d$.  In general, tangent spaces are identified with subspaces of the ambient space. In the current setup, as $\mathcal{P}_d$ is an open sub-manifold, we identify the tangent space at $G$ as $T_G \mathcal{P}_d = \mathcal{S}_d$, the space of symmetric matrices of size $d \times d$. 

Then, in order to turn this differential manifold into a Riemannian manifold, we need to equip its tangent spaces with a metric. 
One choice of metric (leading to a complete Riemannian manifold) is the affine-invariant metric, which, for $S_A, S_B \in T_G \mathcal{P}_d$, is defined as:
\begin{align}
\left\langle S_A, S_B \right\rangle_G = \tr \left(G^{-1} S_A G^{-1} S_B \right).
\label{eq:scalarProdTangplan}
\end{align} 

Note that a Riemannian gradient (that is used by first-order Riemannian optimization methods) is defined with respect to a given tangent space. Then, in order to be coherent with the metric of the tangent space (defined in Eq.~\eqref{eq:scalarProdTangplan}), $\text{grad}_G f(G)$, the Riemannian gradient at $G$ can be obtained from the Euclidean gradient $\nabla f$ as
\begin{align*}
\text{grad}_G f(G) &= G \text{sym} (\nabla f (G)) G.
\end{align*}

\begin{figure}[t]
\centering
   \includegraphics[width=0.65\linewidth]{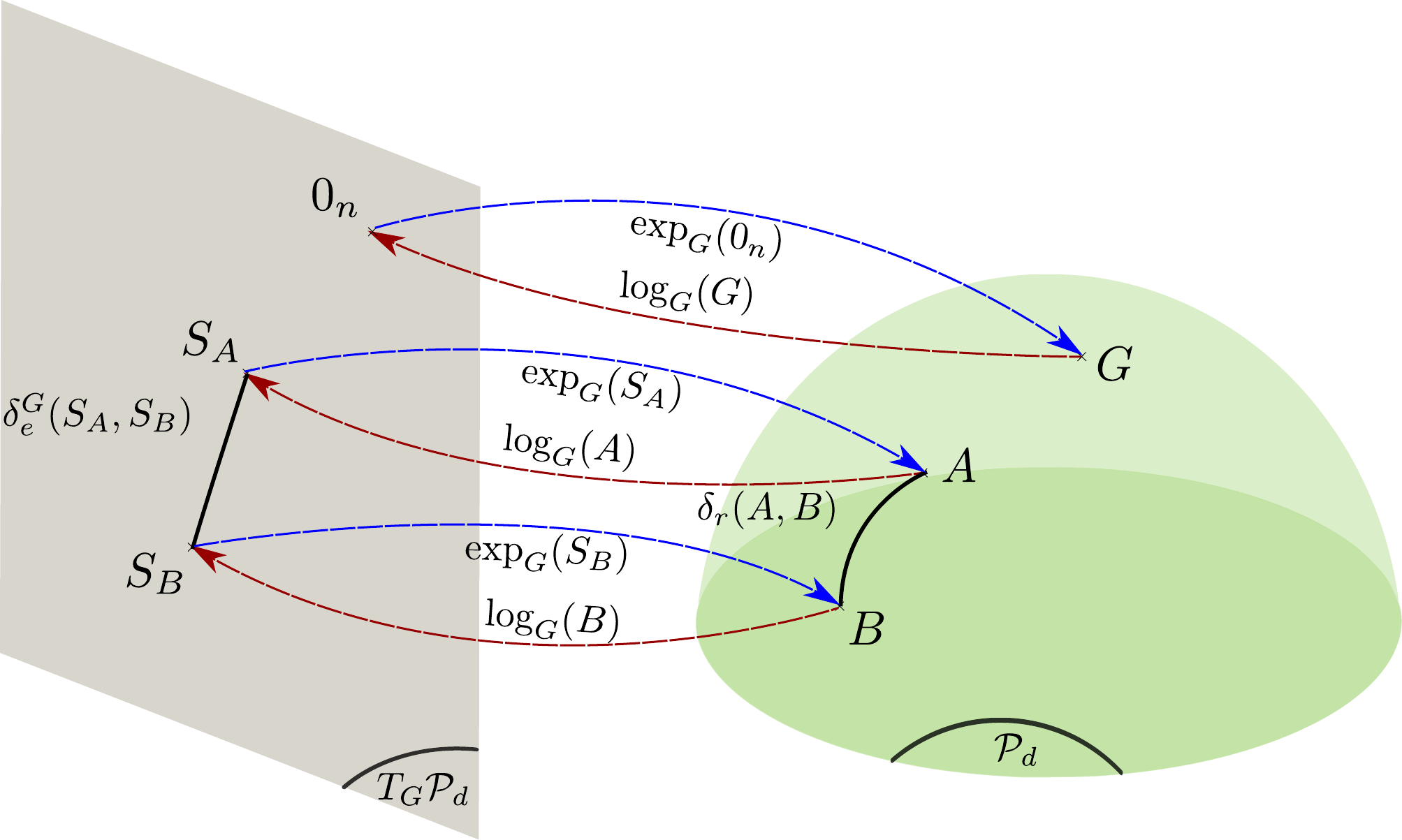}
\caption{Mappings between the manifold of positive definite matrices and the tangent plane at the point $G$. As stated in Th.~\ref{thm:EMIcorolary} $\delta_e^G$, the Euclidean distance in $T_G \mathcal{P}_d$ and the Riemannian distance $\delta_r$ are related but not equal in general.}
\label{fig:EMI}
\end{figure}

As detailed in Appendix~\ref{sec:directionnalderivative},
the Euclidean gradient $\nabla f \left( G \right)$ can be obtained as
\begin{align*}
\nabla f \left( G \right) &=\sum_{ij}Z_{ij}(G) \nabla h_{ij}\left(G  \right),
\end{align*}
where
\begin{align*}
Z(G)&= U \left( \frac{yy\trans}{\norm{U h (G) U}} - \frac{f\left(G\right) U h\left(G\right) U}{\norm{U h(G) U}^2} \right) U,\\
\nabla h_{ij}\left(G  \right) 
&= X_i^{-\frac{1}{2}} \left( DQ_i(G) \left[ \text{sym} \left(A_{ij}\right) \right] \right) X_i^{-\frac{1}{2}} \\
&\phantom{=}
+ X_j^{-\frac{1}{2}} \left( D Q_j(G) \left[ \text{sym} \left(A_{ji}\right) \right] \right) X_j^{-\frac{1}{2}},
\\
Q_i(G)&=\log \left(X_i^{-\frac{1}{2}} G X_i^{-\frac{1}{2}} \right),\\
A_{ij}&=X_i^{\frac{1}{2}}X_j^{-\frac{1}{2}}Q_j(G)  X_j^{\frac{1}{2}}X_i^{-\frac{1}{2}},\\
\text{sym} \left(A \right) &=\frac{A + A\trans}{2},
\end{align*}
where
$Df \left( G \right) \left[  \dot{G} \right]$ is the \emph{directional derivative}
of $f \left( G \right)$ in the direction $\dot{G}$.
To our knowledge, there is no closed-form formula for
computing the directional derivatives
$DQ_i(G) \left[ \text{sym} \left(A_{ij}\right) \right]$ and
$DQ_j(G) \left[ \text{sym} \left(A_{ji}\right) \right]$.
However, we can \emph{numerically} evaluate them,
as shown in~\citet{boumal2011discrete} and \citet{al2009computing}.

Then any displacement in a tangent space $T_G \mathcal{P}_d$ can be mapped back on the manifold (and vice versa) using the reciprocal logarithmic and exponential mappings (see Fig.~\ref{fig:EMI}). 
%To benefit from the Euclidean spaces (tangent spaces) defined on every point of our manifold, we need some operations to map points from the manifold to a tangent space and vice versa. It is done by the reciprocal logarithmic and exponential mappings (see Fig.~\ref{fig:EMI}). 
Any symmetric matrix $S_A$ belonging to $T_G \mathcal{P}_d$, the tangent space at $G$, can be mapped on $\mathcal{P}_d$ (with the reciprocal operation) as
\begin{align}
A =& \exp_G \left( S_A \right) = G^{\frac{1}{2}} \exp \left( G^{-\frac{1}{2}} S_A G^{-\frac{1}{2}}\right) G^{\frac{1}{2}},
\label{eq:expMapping}\\
%\end{align}
%and conversely:
%\begin{align}
S_A = &\log_G \left( A \right) = G^{\frac{1}{2}} \log \left( G^{-\frac{1}{2}} A G^{-\frac{1}{2}}\right) G^{\frac{1}{2}}.
\label{eq:logMapping}
\end{align}

\begin{figure}[t]
\begin{center}
   \includegraphics[width=0.65\linewidth]{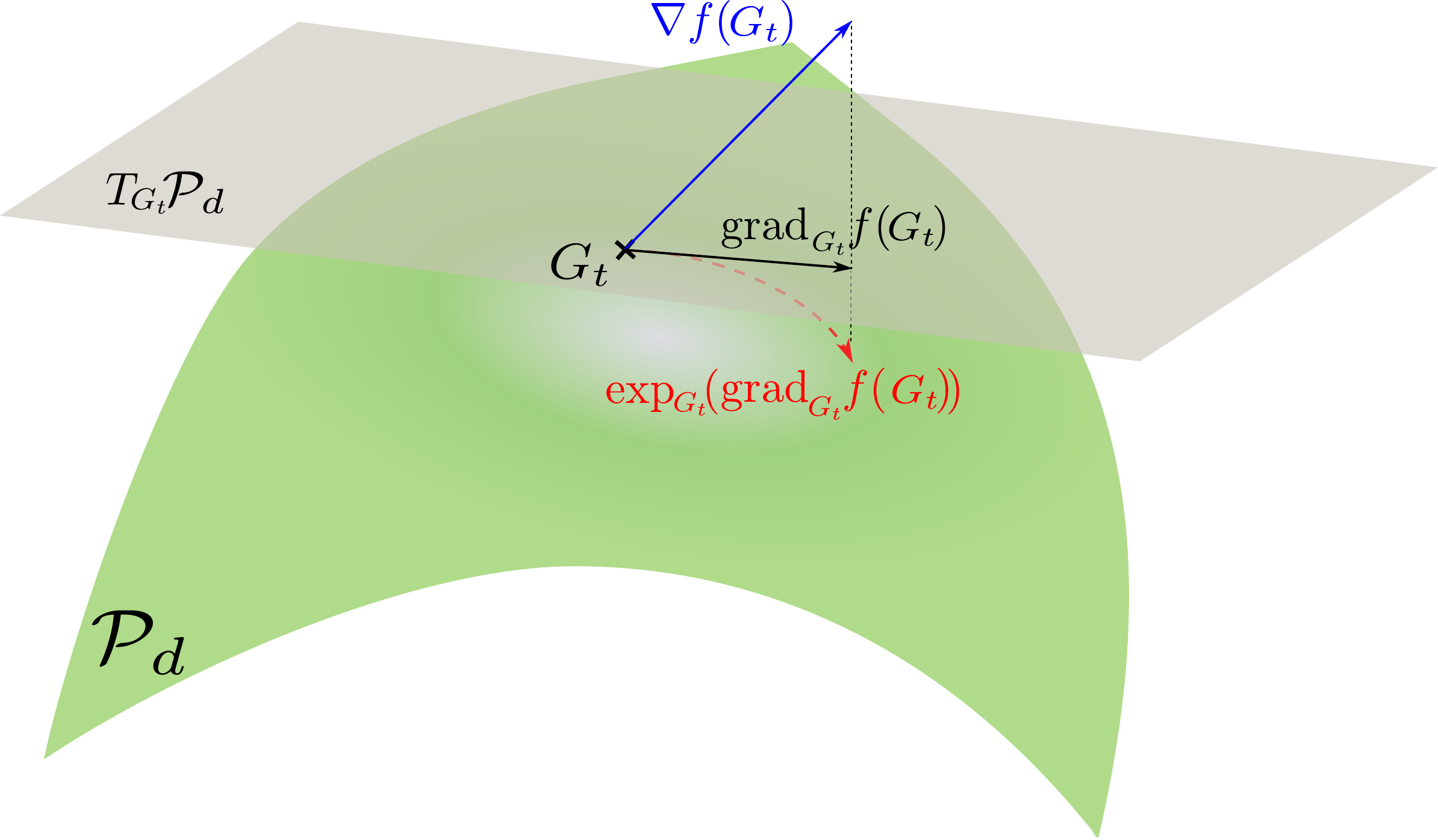}
\caption{Graphical summary of a geodesic gradient descent method for a function $f$ on $\mathcal{P}_d$. }
\label{fig:optiInOneFig}
\end{center}
\end{figure}

These formula give the basic ingredients for implementing a geodesic gradient ascent method (as illustrated in Fig.~\ref{fig:optiInOneFig}) for our objective function\footnote{More details about optimization on Riemannian matrix manifolds
and the geometry of $\mathcal{P}_d$ 
can be found in~\citet{absil2009optimization} and in~\citet{bhatia2009positive}.}.
Indeed, from Eq.~\eqref{eq:expMapping}, the geodesic is given by
\begin{align*}
G_{t+1} =& \exp_{G_t} \left( \eta \text{grad}_{G_t} f(G_t) \right),
\end{align*}
which leads to Eq.~\eqref{geodesic}.

\section{Discussions}
\label{sec:discussion}
In this section, we motivate the use of the LogEuclidean distance and compare it to other distances in $\mathcal{P}_d$.
\subsection{Geometries on $\mathcal{P}_d$}
Depending on the type of geometry chosen, there exist several choices of distances and divergences for comparing SPD matrices. As already discussed in literature~\cite{arsigny2006log, pennec2006riemannian, dryden2009non, fletcher2004principal, cherian2011efficient}, different tools come along with various implicit invariance properties and computational complexities.

As explained in~\citet{bhatia2009positive}, when using the geometry implied by the metric in Eq.~\ref{eq:scalarProdTangplan}, the distance between two SPD matrices $A$ and $B$
is computed along a curve (called the geodesic) and the associated
\emph{affine-invariant Riemannian metric} (AIRM) distance is defined as
\begin{align}
 \delta_r (A,B)& = \norm{ \log( A^{-\frac{1}{2}} B A^{-\frac{1}{2}})}\label{eq:riemDist} \\
 & = \left( \sum_{i=1}^d \log^2 \lambda_i\left(A,B \right)\right)^{\frac{1}{2}} \notag
\end{align}
where $\lambda_i\left(A,B \right)$ are the eigenvalues of the \emph{pencil} $(A,B)$.

As illustrated in Fig.~\ref{fig:psd2d}, when using the Riemannian geometry, the space $\mathcal{P}_d$ becomes a complete manifold. As already stated in~\citet{arsigny2007geometric}, this distance is immune to the swelling effect.
Thus, it could be a good candidate for distance metric learning for covariance matrices. 

However, the AIRM distance comes along with an invariance to a class of congruent transforms~\cite{bhatia2009positive} and it leads to the following isometry for any $M \in \mathcal{P}_d$:
\begin{align}
\delta_r (\Gamma_M(A),\Gamma_M(B))& =\delta_r (A,B).
\label{eq:isometry}
\end{align}
Hence, unlike the LogEuclidean distance, the AIRM distance is immune to the transform $\Gamma(.)$ and it can not be learned using our approach.

Apart from their difference in terms of invariance properties, it should be highlighted that the AIRM distance is not negative-definite and then, contrary to the LogEuclidean distance, cannot be used for defining positive-definite kernels on SPD matrices~\cite{haasdonk2004learning, sra2011positive, barachant2013classification}.

Using information geometry and extending divergences to the matrix case~\cite{cherian2011efficient,sra2011positive,sra2012new}, a symmetrized LogDeterminant divergence (also called the symmetrized Stein loss) can also be used. This divergence can be seen as an approximation of the AIRM distance and as such, there exist some bounds between the AIRM distance and this symmetrized divergence~\cite{sra2011positive}. Moreover, this divergence is invariant to the same transformation as the AIRM distance, but can lead under some conditions to a definite-positive kernel.

In this work, we look for a distance on SPD matrices tailored for classification.
Due to the non-completeness of the space and the swelling effect, the Euclidean distance is not a valid candidate. Hence, the distances derived from Riemannian geometries (and their approximations) are more promising candidates, but in some cases, they are difficult to parametrize. Using the congruent transform $\Gamma_{G^{-\frac{1}{2}}}(.)$, it is possible to parametrize the LogEuclidean distance but not the AIRM distance nor the symmetrized Stein loss due to their invariance properties. 

\begin{figure*}[t]
\centering
   \includegraphics[width=0.95\linewidth]{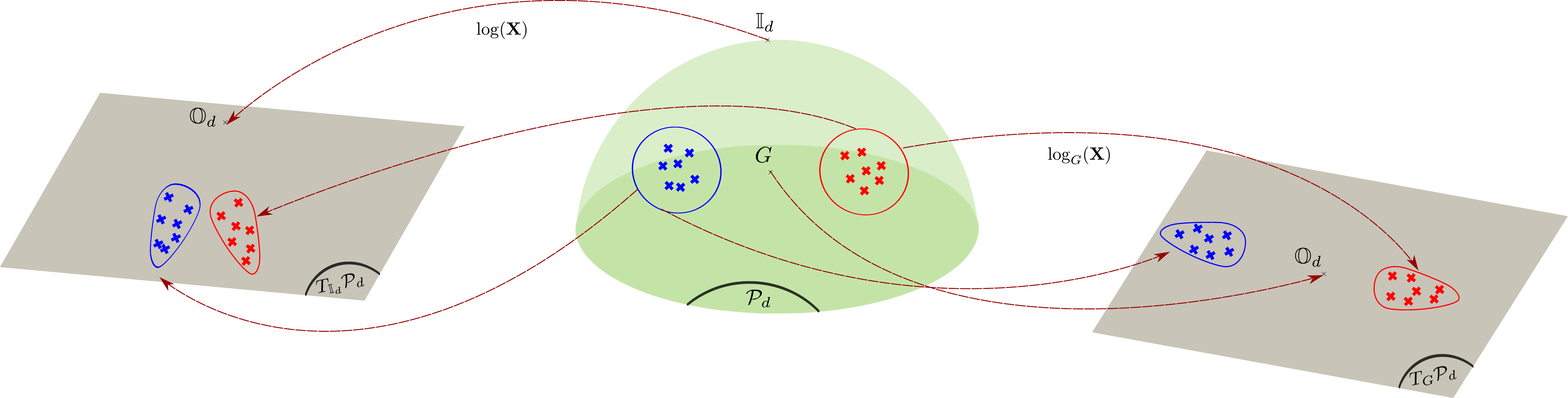}
\caption{Illustration of the deformations implied by two mappings, respectively centered at the identity or at a point $G$}
\label{fig:mappingIllust}
\end{figure*}

\subsection{Interpretations of LogEuclidean metric learning}
Compared to the Euclidean and AIRM distances, the LogEuclidean distance has several advantages:
it is a Riemannian distance (immune to the swelling effect) and is easy to compute. 
In fact, using the fact that the mapping in Eq.~\eqref{eq:logMapping} transforms matrices from the curved space $\mathcal{P}_d$ to the flat space $\mathcal{S}_d$, the LogEuclidean distance can be interpreted as the Euclidean distance between the matrices mapped into $\mathcal{S}_d$.

Interestingly, this LogEuclidean metric can be interpreted as the scalar product,
\[
\langle S_i, S_j \rangle_G= \tr \left(G^{-\frac{1}{2}} S_i G^{-1} S_j G^{-\frac{1}{2}}\right),
\]
in the tangent plane of $\mathcal{P}_d$ at the point $G$, with $S_i$ and $S_j$ the mapping around $G$ of $X_i$ and $X_j$ through the logarithmic map. Hence, according to the taxonomy developed in~\citet{bellet2013survey}, since we optimize $G$ through a logarithmic mapping on a Riemannian manifold, our approach is an instance of non-linear similarity learning.

In this view, the parameter $G$ can be seen as the \emph{center} of the tangent plane,
and it has been empirically observed~\cite{yger2013review} that the choice of $G$ has a strong impact on the metric behaviour. So far, $G$ has been heuristically tuned using either the identity matrix or the Riemannian mean~\cite{moakher2005differential,jeuris2012survey} of data.
However, as highlighted later, mapping a manifold onto a tangent space implies a deformation of the data. This deformation can either be harmful or beneficial to classification. By optimizing the choice of $G$ with respect to a discrimination criterion, we try to force the deformation to improve classification performance.
As shown in Eq.~\eqref{eq:scalarProdTangplan}, every tangent space is equipped with a different metric\footnote{The scalar product defined at every tangent space $T_G \mathcal{P}_d$ is close to a Mahalanobis scalar product parametrized by $G^{-1}$.}. Hence, choosing a new reference for a tangent space leads to choosing a new scalar product. In that sense, learning the reference point of a tangent space is equivalent to learning a metric. 

The LogEuclidean metric is simply a scalar product (i.e., a linear kernel) applied to data mapped through a matrix logarithm.
For this reason, it has sometimes been referred to as the \emph{LogEuclidean kernel}~\cite{barachant2013classification,yger2013review, jayasumana2013kernel}. In this regard, the optimization of the LogEuclidean metric can also be interpreted as a kernel learning approach.

The use of the LogEuclidean distance can be interpreted as flattening the Riemannian manifold. As discussed below, such a mapping implies some deformations of the space of SPD matrices (see Fig.~\ref{fig:mappingIllust}). Our approach can then be interpreted as finding the mapping such that the implied deformations benefit classification performance.

\subsection{Geometrical motivation}
The deformation implied by mapping points in a tangent space has been first stated as the \emph{exponential metric increasing} (EMI) property in~\citet{bhatia2009positive}. It gives inequalities between the Riemannian distance $\delta_r$ and the Euclidean distance $\delta_e$ in the tangent space at the identity $\mathbb{I}_d$.
\begin{theorem}\cite{bhatia2009positive}
\label{thm:EMIprop}
For each pair of points $A$, $B$ in $\mathcal{P}_d$, we have
\[
\delta_r (A,B) \geq \delta_l ( A, B)= \|\log(A) - \log(B)\|_{\mathcal{F}}.
\]
\end{theorem}

The equality occurs when $A$, $B$ and $\mathbb{I}_d$ belong to the same geodesic. This property, inherited from the non-positive curvature of $\mathcal{P}_d$, implies a deformation of the shapes when using the logarithmic map.

Using the properties of the bijection $\Gamma_{G^{-\frac{1}{2}}}(.)$ and with the notations in Eq.~\eqref{eq:scalarProdTangplan} and in Eq.~\eqref{eq:logMapping}, a corollary has been proposed in~\citet{yger2013review} which extends Theorem~\ref{thm:EMIprop} to any tangent space:
\begin{theorem}\citep{yger2013review}
\label{thm:EMIcorolary}
For any $A$, $B$ and $G$ in $\mathcal{P}_d$, with $\|\cdot\|_G$,
the norm associated to the natural scalar product in $T_G \mathcal{P}_d $ satisfies
\begin{align*}
%\delta_r (A,B) \geq  \overbrace{\| \log_G(A) - \log_G(B)\|_{G}}^{\delta_e^G (\log_G(A), \log_G(B))} .
\delta_r (A,B) \geq \delta_l^G(A,B)= 
\| \log_G(A) - \log_G(B)\|_{G} .
\end{align*}
\end{theorem}
Similarly to Theorem~\ref{thm:EMIprop},
the equality occurs when $A$, $B$ and $G$ belong to the same geodesic.

From these two theorems, as illustrated in Fig.~\ref{fig:mappingIllust}, it follows that changing the reference point for centering the tangent space modifies how shapes are deformed once they are mapped in this tangent space.
This motivates our metric learning approach.

\section{Numerical experiments}
\label{sec:expeNum}
In this section, we report experimental results.

\subsection{Setup}
Concerning the implementation of our approach,
we employed a \emph{Riemannian trust-region}\footnote{with the implementation provided in the Manopt toolbox provided by~\citet{boumal2013manopt}.}~\cite{boumal2011discrete,absil2009optimization} .
In practice, we use the following regularized optimization problem:
% During our preliminary experiments, we noticed that optimizing the problem in 
% Eq.~\ref{eq:optiPb} can often lead to numerically ill-conditioned iterates\footnote{Note also that the un-normalized KTA criterion is nicer in term of numerical convergence but it showed poor performances that are not reported in the experimental section.}. In order to prevent this, we decided to apply some regularization and then perform the optimization within a certain radius around the initial point $G_0$ (i.e., until we find an iterate $G_t$ violating the constraint $\delta_r (G_0, G_t) \leq \epsilon$):
\[
\begin{cases}
%\displaystyle\max_{G \in \mathcal{P}_d}\mathcal{A}(K_G, yy\trans)\\
\displaystyle\max_{G \in \mathcal{P}_d} f(G)\\
\text{s.t.~}\quad \delta_r (G_0, G) \leq \epsilon
\end{cases}
 \]
for some $G_0$.
As we have not found any guarantee concerning the convexity or the geodesic-convexity~\cite{wiesel2012geodesic} of the centered KTA,
we decided to use the Riemannian mean~\cite{moakher2005differential,jeuris2012survey} as $G_0$.
Below, we set $\epsilon=10$ for all datasets.

In our numerical experiments, we report accuracies on balanced datasets using a 1-nearest neighbor (1-NN) classifier equipped with different distances. Our baseline distances for the 1-NN classifier comprise the Euclidean distance $\delta_e$, the Riemannian distance $\delta_r$ and the LogEuclidean distance (parameterized by the identity matrix or the Riemannian mean of the training samples). We compare all these baseline distances to the LogEuclidean distance $\delta_l^G$ with parameter $G$ learned by our approach. In this section, when the reported results are averaged over several iterations (as in Tab.~\ref{tab:toyExpe} and Tab.~\ref{tab:textureExpe}), we show when the improvement is significant\footnote{by comparing the two best results with the two-sided Wilcoxon signed rank test at significance level $0.05$.} using a bold font.

\subsection{Toy dataset}
First, in order to gain some insights on the behavior of our approach, we designed a simple experiment in which, the covariance matrices are generated as follows:
\begin{align*}
X &= Q\text{diag} (\lambda_1, \ldots, \lambda_r,\mu_1, \ldots, \mu_r ) Q^\top \\
&\phantom{=}+ V\text{diag} (|\epsilon_1|, \ldots, |\epsilon_{2r}| ) V^\top
\end{align*}
with $Q$ and $V$, two random orthonormal square matrices, 
and $\lambda_i \sim \mathcal{N}(5,0.2)$ or $\lambda_i \sim \mathcal{N}(4,0.1)$ respectively for the positive and negative classes and $\mu_i \sim \mathcal{U}([1,6])$ and $\epsilon_i \sim \mathcal{N}(0,1)$ independent of the class. 
The dataset is composed of $50$ samples in the training set and $500$ samples in the test set. It is re-sampled in order to repeat the experiment $10$ times and the results are reported in Tab.~\ref{tab:toyExpe}.

\begin{table}[t]
\caption{Comparison of the mean accuracy of the LogEuclidean distance (parameterized by our method $G_{\text{KTA}}$) to other distances on a toy dataset with growing sizes of covariance matrices.}
\label{tab:toyExpe}
\centering
\begin{tabular}{|c||c|c|c||c|c|}
\hline
& \multicolumn{3}{c||}{LogEuclidean} & & \\\cline{2-4} 
size	 & $\mathbb{I}_d$ & $\bar{X}$ &	$G_{\text{KTA}}	$ & 	$\delta_r$& 	$\delta_e$ \\\hline\hline
6x6	&77.50&	77.58&		\textbf{84.78}	&77.66	&82.76\\\hline
8x8&	77.46&	77.44&		\textbf{86.78}&	77.42&	83.32\\\hline
16x16&	75.58&	75.62&	\textbf{89.56}&	75.48&	85.78\\\hline
20x20&	77.06&	77.10	&	\textbf{90.96}&	77.16&	87.16 \\\hline
\end{tabular}
\end{table}

In this simple experiment, as it was foreseeable, the Euclidean distance achieves good performances. Indeed, omitting the additive noise matrix, the data generation protocol only involves a fixed orthonormal matrix $Q$ for generating the matrices from both classes. Hence, the separation between the classes just becomes linear in terms of the eigenvalues of the covariance matrices. Interestingly, the Riemannian distances (except our approach) performs badly for all cases.
% even when the dimension of the problem grows.

From this experiment, it is interesting to note that the proposed metric learning method helps the LogEuclidean distance to overcome its limitation.. Hence, the LogEuclidean distance with parameter learned by our proposed method is the best not only among the Riemannian distances but among all candidates. However, this toy experiment may be too simple for our applications and we test our methods on real-life datasets below.

\subsection{Brain Computer-Interface}
We tested our approach on Dataset 2a from the BCI Competition IV\footnote{\url{http://www.bbci.de/competition/iv/}}~\cite{naeem2006seperability}. 
The dataset consists of EEG signals (recorded from $22$ electrodes) on $9$ subjects who were asked to perform left hand, right hand, foot and tongue motor imagery~\cite{pfurtscheller2001motor}
(i.e., EEG signals are recorded while the subject is only imagining the given limb movements without actually moving it). 

Following the protocol described in~\cite{lotte2011regularizing}, we only used the EEG signals corresponding to the left and right hand.
Hence, for every subject, we have $72$ trials (i.e. segments of signal) for each class in both training and testing.
Moreover, we applied the same band-pass filtering ($[8-30]$ Hz) in order to pre-process the raw signals.

As shown in literature~\cite{barachant2010riemannian,barachant2013classification,yger2013review}, for motor imagery signals, the spatial covariance matrix of the trials is a very promising feature. 
Based on this observation, we extracted the covariance matrix between sensors (i.e., the spatial covariance matrix) from every filtered time segment (i.e., every example to be classified).

\begin{figure*}[t]
\centering
   \includegraphics[width=0.8\linewidth]{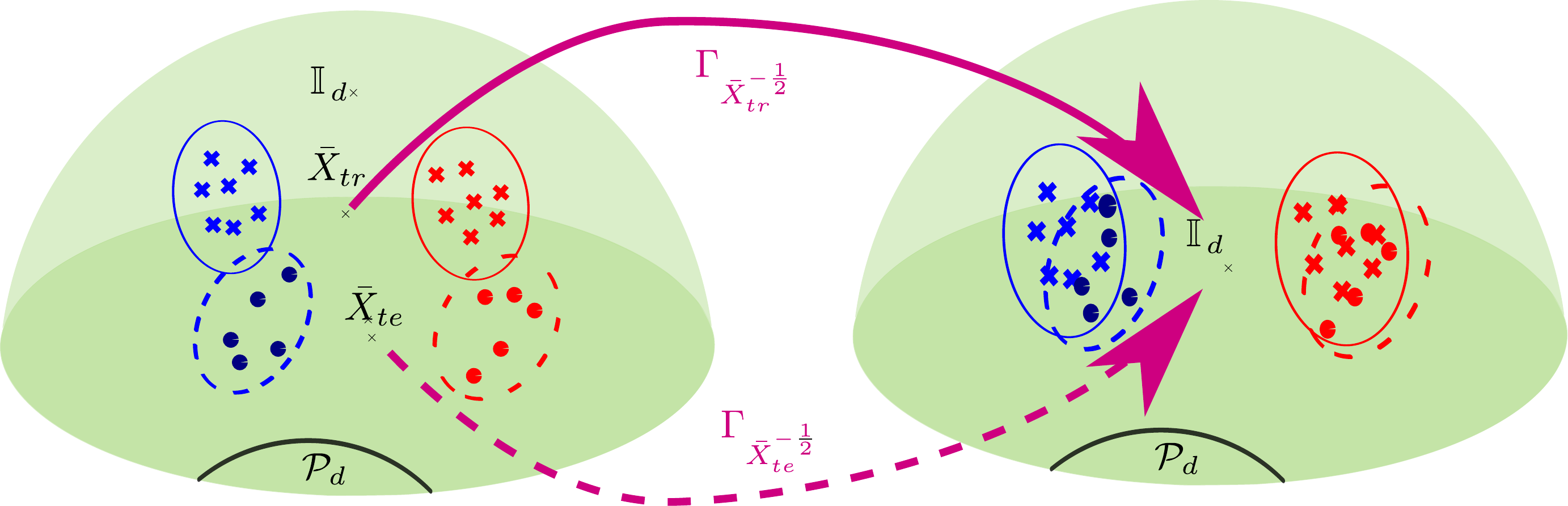}
\caption{Illustration of the effect of whitening when the Riemannian mean of the data is non-stationary.}
\label{fig:preproc}
\end{figure*}

Although we could directly use our approach and try to classify samples,
the performance would be poor since this dataset is known to show some non-stationarity
between sessions. In order to cope with this problem,  we employ an adaptive kernel formulation
proposed in~\citet{barachant2013classification}. This method cancels
the non-stationarity in the data by whitening the covariance matrices
in the training session and test session by subtracting the Riemannian mean 
of the training and test datasets, respectively.
In practice, whitening is carried out by first applying the congruent transformation
$\Gamma_{\bar{X}^{-\frac{1}{2}}}(\cdot)$, where $\bar{X}$ is the Riemannian mean of the data set.
% , for every matrix $X_i$, by left- and right-multiplying it by $\bar{X}^{-\frac{1}{2}}$ (i.e. a congruent transformation), the Riemannian mean of the set. 
As illustrated in Fig.~\ref{fig:preproc}, this whitening process has the effect of spreading the covariance matrices from both training and test dataset around the identity matrix.
For a dataset spread around $\bar{X}$, this has the effect of ``transporting'' the data on the manifold along the geodesic. For real-life applications, this whitening process can be criticized since it uses the test data. However, this can be carried out in a semi-supervised fashion
by estimating $\bar{X}^{-\frac{1}{2}}$ during the calibration phase. 
Therefore, this process would still be practical in many applications.

As reported in Tab.~\ref{tab:BCIExpe}, compared to other distances for
handling covariance matrices, our proposed approach performs the best on 
$5$ subjects and is never the worst. On average over the subjects,
we observe a significant gain by using an optimized reference point
for the LogEuclidean distance.

\begin{table}[t]
\centering
\caption{Comparison of the accuracy of the LogEuclidean distance (parameterized by our method $G_{\text{KTA}}$) to other distances for the classification of hand movements for $9$ subjects of the BCI Competition IV Dataset 2a.}
\label{tab:BCIExpe}
\begin{tabular}{|c||c|c||c|c|}
\hline
& \multicolumn{2}{c||}{LogEuclidean} & &\\\cline{2-3}
subject & $\mathbb{I}_d$ & 	$G_{\text{KTA}}	$ & 	$\delta_r$ &	$\delta_e$ \\\hline
1	&	80.56	& 	83.33	&	79.17	&77.08\\\hline
2	&	58.33	& 	54.17	&	59.03	&52.78\\\hline
3	&	93.06	& 	95.83	&	93.75	&88.19\\\hline
4	&	54.17	& 	54.17	&	52.78	&56.25\\\hline
5	&	52.78	& 	55.56	&	52.78	&53.47\\\hline
6	&	63.89	& 	61.81	&	62.50	&60.42\\\hline
7	&	52.78	& 	60.42	&	52.78	&61.11\\\hline
8	&	93.75	& 	97.92	&	92.36	&92.36\\\hline
9	&	83.33	& 	84.72	&	82.64	&84.72\\\hline\hline
mean	&	70.29	& 	71.99	&	69.75	&69.60\\\hline
\end{tabular}
\end{table}

\subsection{Texture classification}
\begin{figure}[t]
\centering
\begin{subfigure}[D2]{
\includegraphics[scale=0.16]{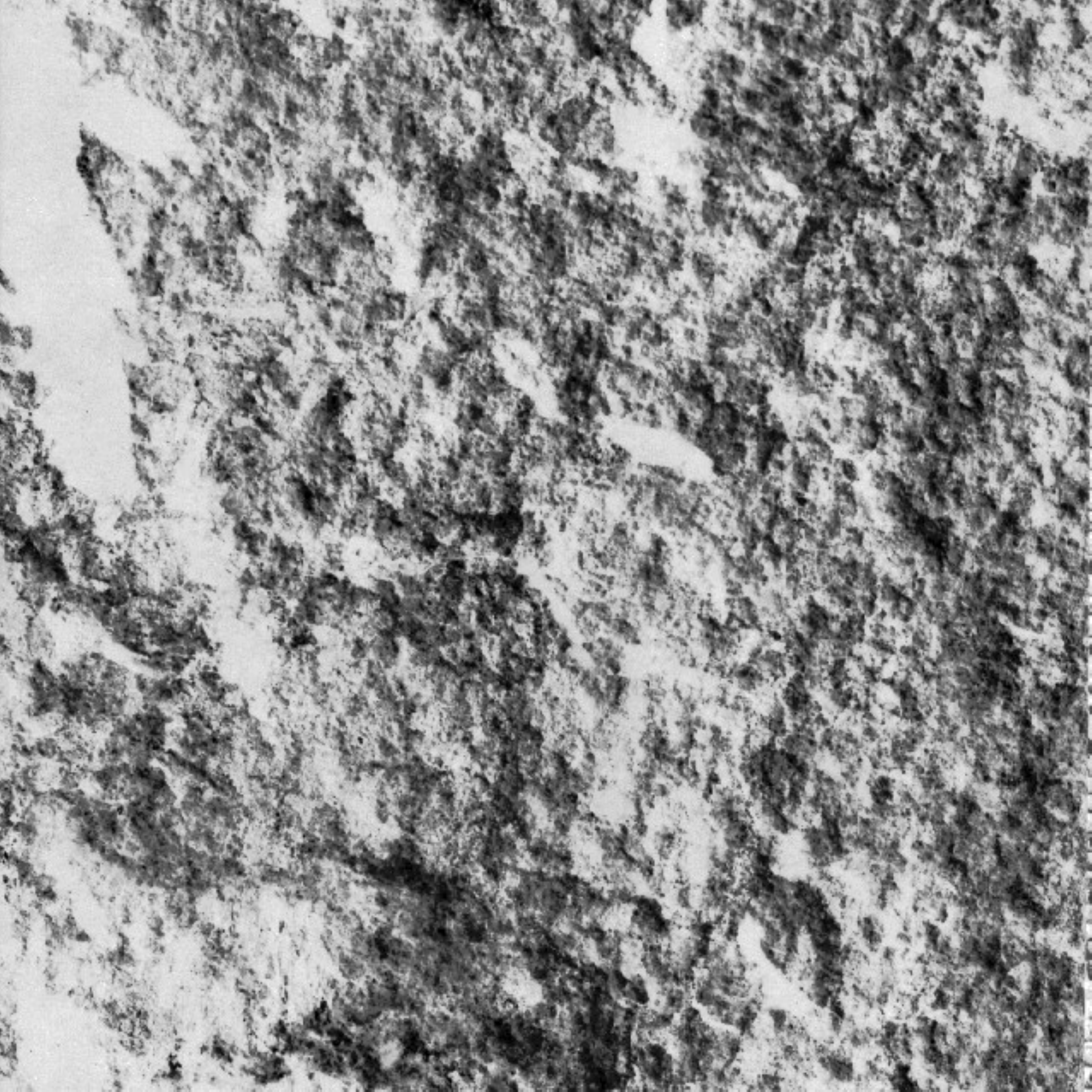}}
\end{subfigure} 
~~
\begin{subfigure}[D28]{
\includegraphics[scale=0.16]{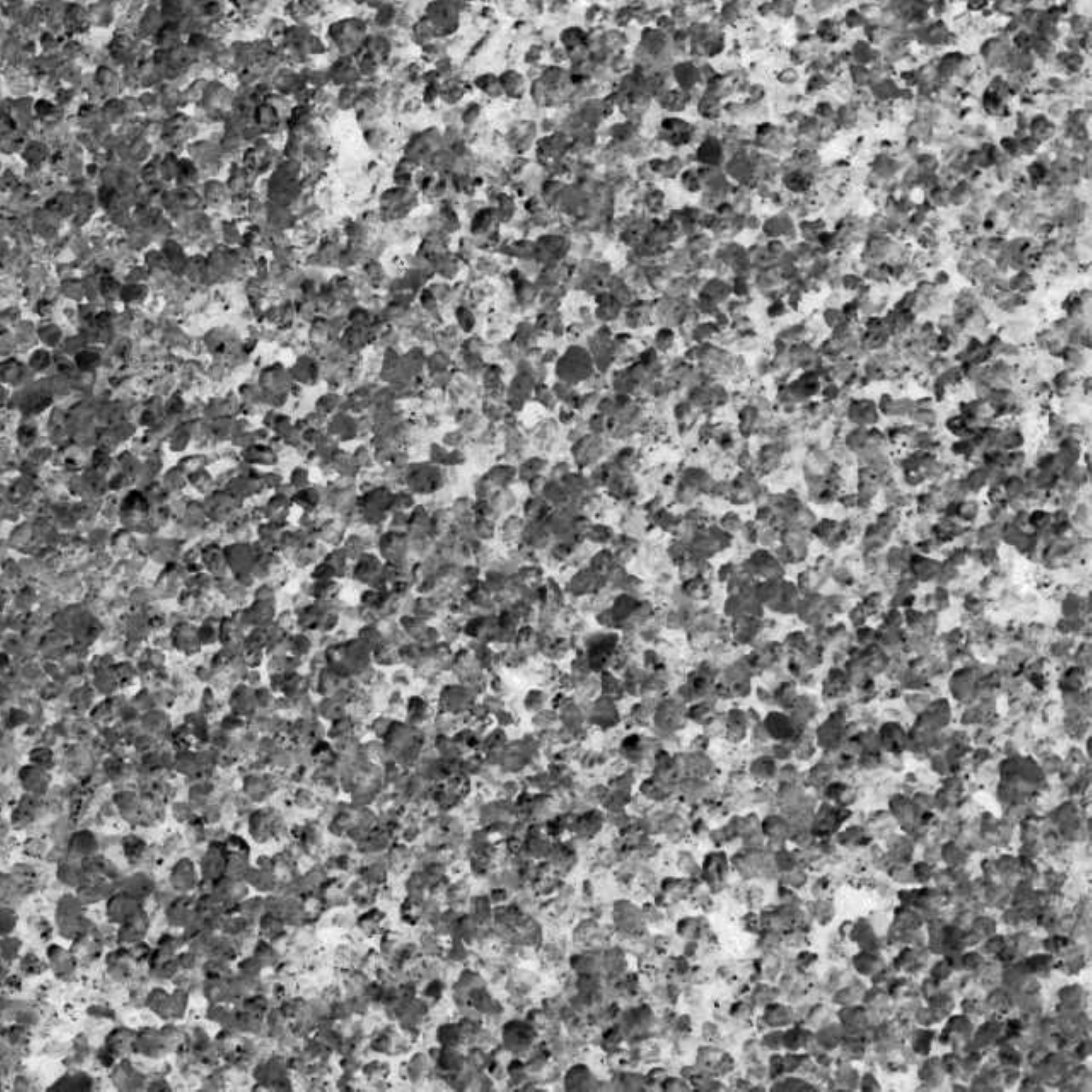}}
\end{subfigure} 
~\\~\\
\begin{subfigure}[D29]{
\includegraphics[scale=0.16]{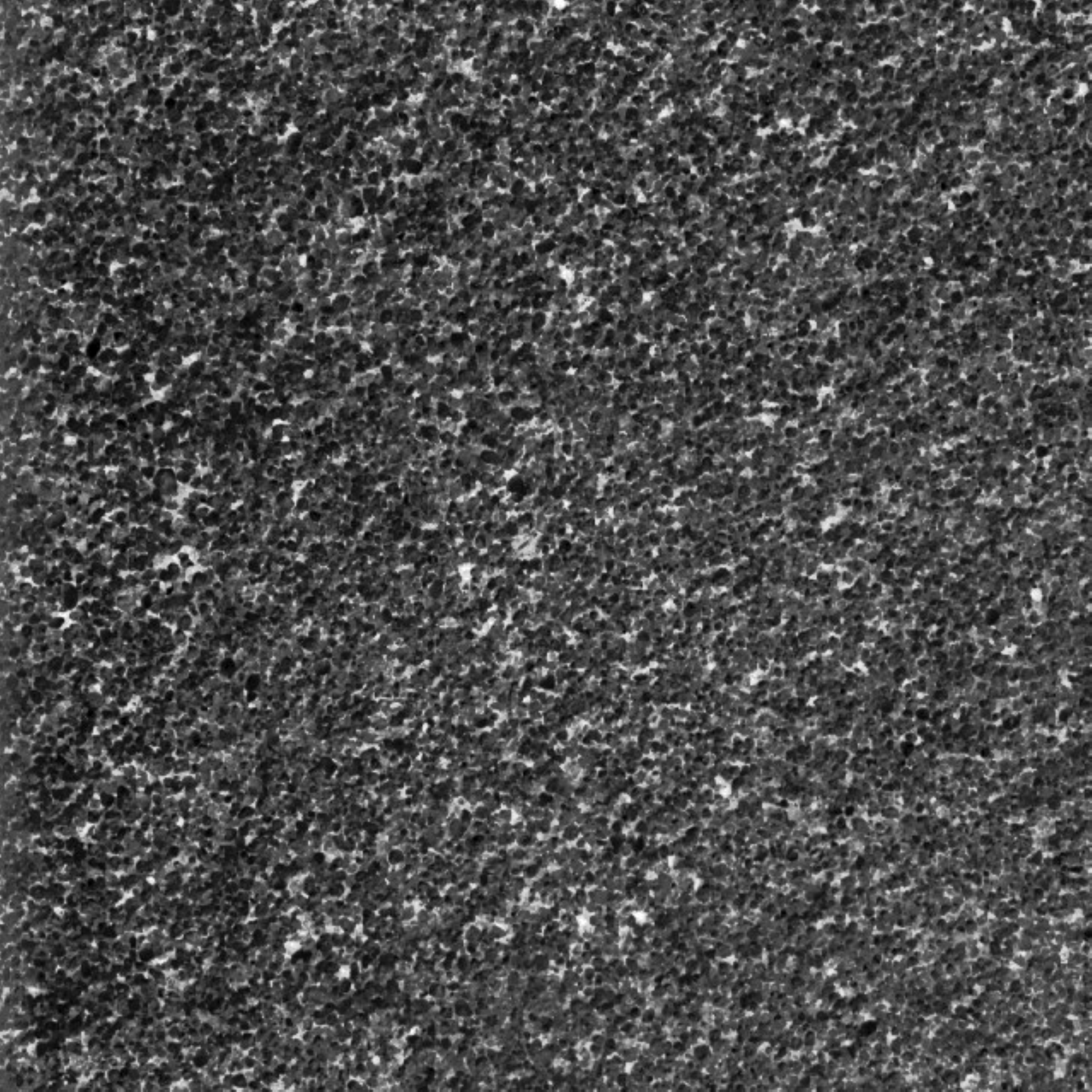}}
\end{subfigure}
~~
\begin{subfigure}[D92]{
\includegraphics[scale=0.16]{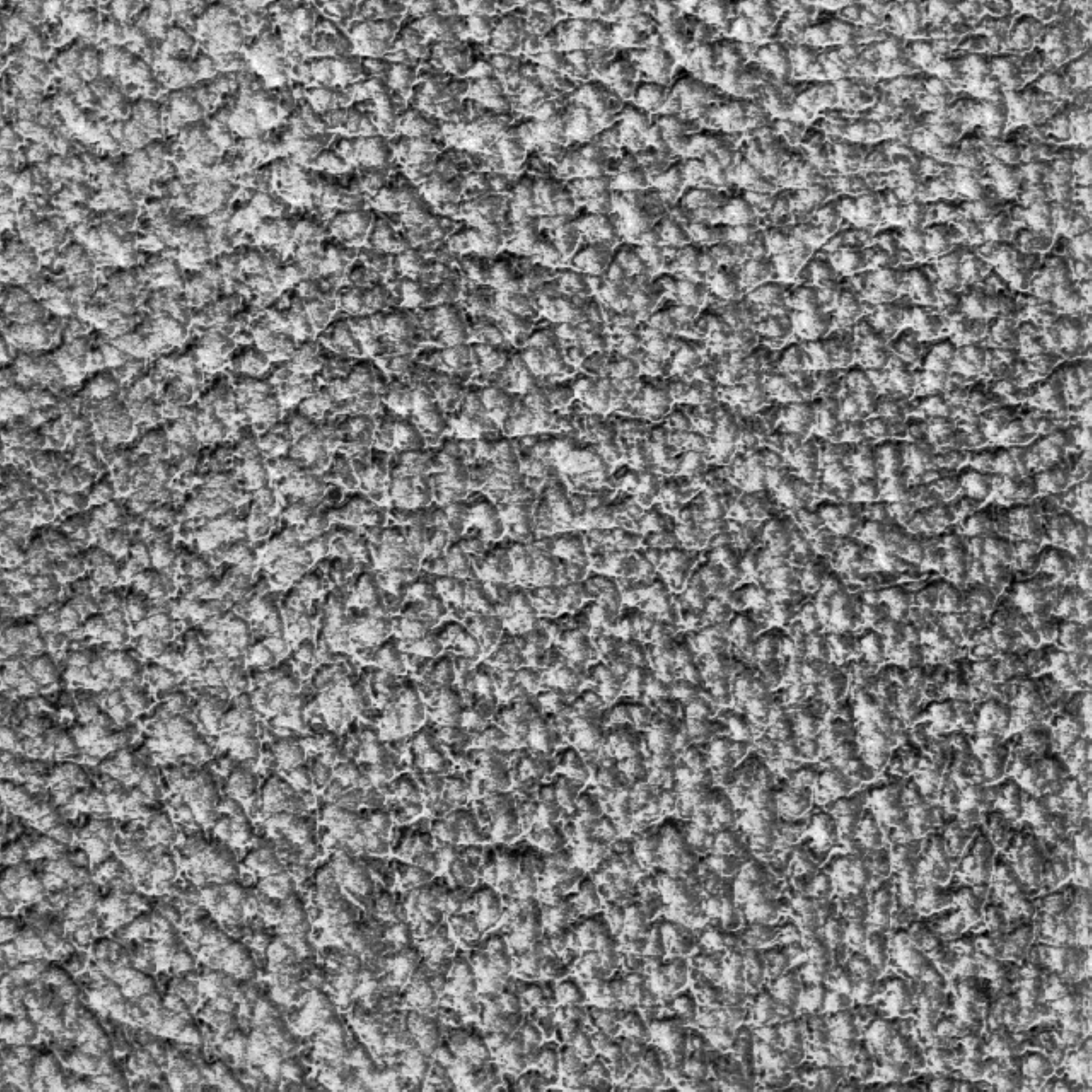}}
\end{subfigure}
\caption{Selected textures from the Brodatz dataset}
\label{fig:BrodatzFig}
\end{figure}

As another example of real-world problems, we consider a subset of four textures (shown in Fig~\ref{fig:BrodatzFig}) of the \emph{Brodatz dataset} \cite{brodatz1966textures}.
Similarly to the works of~\citet{mairal2009supervised} and \citet{yger2011wavelet},
we extracted $16\times 16$ patches from every texture.
For every couple of textures, the training set composes patches from the left half
of each texture and the test set composes patches from the right half.
In every set, patches may overlap, but the training and test sets do not overlap each other. 

Every method was trained on $50$ patches selected from the training set and tested on $300$ patches selected from the test set. Classification rates have been computed as the average of $50$ runs after re-sampling of the training and test sets. 

As proposed in~\citet{tou2009gabor}, we built covariance matrices using a bank of $8$ Gabor filters ($4$ angles and $2$ scales). Hence, each $16\times 16$ patch is transformed into a $8\times 8$ covariance matrix.      
        
\begin{table}[t]
\centering
\caption{Comparison of the mean accuracy of the LogEuclidean distance (parameterized by our method $G_{\text{KTA}}$) to other distances for classifying patches of texture taken from the Brodatz dataset.}
\label{tab:textureExpe}
\begin{tabular}{|c||c|c|c||c|c|}
\hline
& \multicolumn{3}{c||}{LogEuclidean} &  &\\\cline{2-4}
textures	 & $\mathbb{I}_d$ & $\bar{X}$ &	 $G_{\text{KTA}}	$ & 	$\delta_r$&$\delta_e$\\\hline
%D2 vs D28	&	86.57	&	86.67	&	88.90	&	86.43	&63.77\\\hline
%D2 vs D29	&	74.60	&	74.57	&	75.10	&	74.40	&63.97\\\hline
%D2 vs D92	&	74.70	&	74.77	&	74.80	&	74.17	&68.77\\\hline
%D28 vs D29	&	75.83	&	77.07	&	77.53	&	76.50	&55.67\\\hline
%D28 vs D92	&	97.87	&	97.80	&	98.40	&	97.67	&72.93\\\hline
%D29 vs D92	&	85.20	&	85.33	&	88.27	&	85.27	&66.77\\\hline\hline
%mean & 82.46 & 82.7 & 83.83& 82.41 &65.31\\\hline
D2 vs D28	&86.37&	86.58	&	\textbf{88.39}&	86.31&	63.53
 \\\hline
D2 vs D29	&75.5&	75.33	&	\textbf{76.69}	&75.14	&64.03
 \\\hline
D2 vs D92	& 74.88&	74.68&		\textbf{76.71}	&74.33	&68.05
\\\hline
D28 vs D29	&	75.5	0 &75.33		&\textbf{76.69}	&75.14	&64.03
 \\\hline
D28 vs D92	&	74.88	&74.68	&	\textbf{76.71}&	74.33	&68.05
 \\\hline
D29 vs D92	&	83.68	&83.53	&	\textbf{85.83}	&83.48	&65.93
 \\\hline\hline
mean & 78.47	 &78.36 &	80.17&	78.12	& 65.60
\\\hline
\end{tabular}
\end{table}    
        
Tab.~\ref{tab:textureExpe} summarizes the experiments on texture classification. First, it clearly shows the interest of using the Riemannian geometry. Indeed, for this data, the Euclidean distance performs poorly. Moreover, this experiment shows again that our supervised metric learning algorithm for choosing the reference of a LogEuclidean distance is effective.
        
\section{Conclusion and perspectives}
\label{sec:conclusion}
% speed up using stochastic gradient algorithm
% extension to exponential kernel optimization (cf Ramona et papier Harandi)
% multiclasse probleme ?]
% low rank ? (cf from manifold to manifold)
%
In this paper,
we have introduced a novel approach for selecting a LogEuclidean metric.
This approach is founded on theoretical observations about the distortion
implied by a LogEuclidean metric. 
Then, we proposed to cast the problem of choosing a relevant metric as
a kernel learning algorithm with a kernel target alignment criterion.
We finally solved this problem using tools from the field of optimization
on manifold and applied it to synthetic and real-world data.

Although we restricted ourselves to binary classification problems,
the extension of our approach to multi-class problems ~\cite{ramona2012multiclass}
is straightforward.

When huge datasets are involved,
our optimization algorithm might not scale well.
A stochastic setting on manifold~\cite{bonnabel2013stochastic}
could be a promising extention.

We only considered full-rank covariance matrices,
but the extension of our approach to low-rank matrices
is a challenging and interesting future work,
since this corresponds to 
supervised feature extraction under the LogEuclidean metric.
 %As the number of sources (for BCI data) or local feature (for images) grows, the number of feature encoded by a covariance matrix grows quadratically. In that case, 
Exploration along this line of research would bridge the gap between 
our approach and the one proposed in~\citet{harandi2014manifold}. 
%\section*{Acknowledgments} 
% thanks to Pr. Rakoto, Pr. Gasso and Dr. Berar (starting point)
%            Pr. Hartley (for the end), Dr. Boumal (for discussions and help)
% Funding sources -> JSPS (ANR sur le debut du projet ?)
%                 -> Kakenhi
 
% TODO last 
 
%\textbf{Do not} include acknowledgements in the initial version of
%the paper submitted for blind review.
%If a paper is accepted, the final camera-ready version can (and probably should) include acknowledgements. In this case, please place such acknowledgements in an unnumbered section at the end of the paper. Typically, this will include thanks to reviewers who gave useful comments, to colleagues who contributed to the ideas, and to funding agencies and corporate sponsors that provided financial support.  

\bibliography{ref}
\bibliographystyle{icml2015}
%\clearpage
%\newpage
\appendix	
\section{Appendix}
In the appendix, we give more details about the derivation of the our cost function $f$ and we give the results of an extra numerical experiment on a toy dataset.

\subsection{Gradient of f(G)}
\label{sec:directionnalderivative}
Deriving the cost function $f$ using the classical tricks of the trade of matrix derivation is not obvious (if not impossible). Indeed, this function is based on the matrix logarithm whose derivative is not simple to obtain.

To obtain the Euclidean gradient $\nabla f$ of the function $f$, we need $Df (X)[H]$, its directional derivative\footnote{Note that it is sometimes called Fr\'{e}chet derivative.} 
at a point $X$ and in a direction $H$, which is defined as:
\begin{align*}
Df (X)[H] &= \lim_{h \to 0} \frac{f(X + h H) - f(X)}{h}
\end{align*}

In order to obtain the gradient of $f$, we express its directional derivative.
As the the gradient and the directional derivatives are linked with the following equality
\begin{align*}
Df \left( G \right) \left[  \dot{G} \right] &= \left\langle \nabla f \left( G \right), \dot{G} \right\rangle
\end{align*}
we need the adjoint of the directional derivative in order to obtain $\nabla f \left( G \right)$.

Using standard properties of the directional derivatives, we formulate $Df \left( G \right) \left[  \dot{G} \right]$ as :\\~\\~\\~\\~
%\begin{widetext}
\begin{align}
Df \left( G \right) \left[  \dot{G} \right]
&=\frac{\scalProd{U Dh \left( G \right) \left[  \dot{G} \right]U,}{yy\trans} \norm{U h\left( G \right) U} - \scalProd{U h \left( G \right) U,}{yy\trans}    
\scalProd{ \frac{U h \left(G \right) U}{\norm{U h\left( G \right)U}}}{U Dh \left( G \right) \left[  \dot{G} \right] U}
}{ \norm{U h\left( G \right) U}^2}\\
=& \left\langle  Dh \left( G \right) \left[  \dot{G} \right], \underbrace{U \left( \frac{yy\trans}{\norm{U h (G) U}} - \frac{f\left(G\right) U h\left(G\right) U}{\norm{U h(G) U}^2} \right) U}_{Z \left( G \right)} \right\rangle_{\mathcal{F}}
\label{eq:dirDerivF}
\end{align}
%\end{widetext}

In Eq.~\ref{eq:dirDerivF}, the core quantity to estimate is $Dh \left( G \right)$. Hopefully, the function $h \left( G \right)$ has already been studied in~\cite{boumal2011discrete}.
Using those results with the same convention ($\text{sym} \left(A \right) = \frac{A + A\trans}{2} $), we have the gradient of $h_{ij}(G)$:
%\begin{widetext}
\begin{align}
\nabla h_{ij}\left(G  \right) \label{eq:nablah}
=& X_i^{-\frac{1}{2}} \left( D\log \left(X_i^{-\frac{1}{2}} G X_i^{-\frac{1}{2}} \right) \left[ \text{sym} \left(X_i^{\frac{1}{2}}X_j^{-\frac{1}{2}} \log \left( X_j^{-\frac{1}{2}} G X_j^{-\frac{1}{2}}\right)   X_j^{\frac{1}{2}}X_i^{-\frac{1}{2}}\right) \right] \right) X_i^{-\frac{1}{2}} \\
&+ X_j^{-\frac{1}{2}} \left( D\log \left(X_j^{-\frac{1}{2}} G X_j^{-\frac{1}{2}} \right) \left[ \text{sym} \left(X_j^{-\frac{1}{2}}X_i^{\frac{1}{2}} \log \left( X_i^{-\frac{1}{2}} G X_i^{-\frac{1}{2}}\right)   X_i^{-\frac{1}{2}}X_j^{\frac{1}{2}}\right) \right] \right) X_j^{-\frac{1}{2}} \notag
\end{align}
%\end{widetext}

Note that the gradient of $\nabla h_{ij}$ depends on the directionnal derivative of the matrix logarithm. To our knowledge, there is no closed-form formula of this directionnal derivative but is can be evaluated numerically using the algorithm in~\citep{boumal2011discrete,al2009computing}.

Then, the Euclidean gradient of $f$ at $G$ is expressed with respect to the adjoint of $Dh \left( G \right)$:
\begin{align}
\nabla f \left( G \right) = \left(Dh \left( G \right) \right)^{\star} \left[ Z \left( G \right)\right].
\end{align}

Hence, after some algebra (detailed in the next section), we express the adjoint as :
\begin{align}
&\left\langle Z, Dh \left(G\right) \left[\dot{G}\right] \right\rangle_{\mathcal{F}}
= \left\langle \overbrace{\sum_{ij}Z_{ij} \nabla h_{ij}\left(G  \right) }^{\left(Dh \left( G \right) \right)^{\star} \left[ Z \left( G \right)\right]}, \dot{G}\right\rangle_{\mathcal{F}}.
\end{align}

To sum up, using Eq.\ref{eq:nablah} and Eq.\ref{eq:dirDerivF}, we have :
\begin{align}
\nabla f \left( G \right) =\sum_{ij}Z_{ij} \nabla h_{ij}\left(G  \right). 
\end{align}

Note that, before solving our optimization problem on data, the validity of our Euclidean gradient formula has been numerically checked using the sanity checking tools provided in the Manopt toolbox by \citet{boumal2013manopt}. 

\subsection{Adjoint of $Dh \left( G \right)$}
In order to obtain the adjoint, using the decomposition over the canonical basis $\dot{G}=\sum_{k,l} \dot{G}_{kl} e_k e_l\trans$, we write :
\begin{align}
&\left\langle Z, Dh \left(G\right) \left[\dot{G}\right] \right\rangle_{\mathcal{F}}\\
=& \sum_{ij}Z_{ij}\left( Dh \left(G\right) \left[\dot{G}\right] \right)_{ij}\\
=& \sum_{ij} \left\langle \nabla h_{ij}\left(G  \right), \dot{G}   \right\rangle_{\mathcal{F}} Z_{ij}\\
=& \sum_{kl}\sum_{ij} \left\langle \nabla h_{ij}\left(G  \right), Z_{ij} e_k e_l\trans\right\rangle_{\mathcal{F}} \dot{G}_{kl}\\
=& \sum_{kl} \left(\sum_{ij}Z_{ij} \nabla h_{ij}\left(G  \right) \right)_{kl} \dot{G}_{kl}\\
=&\left\langle \overbrace{\sum_{ij}Z_{ij} \nabla h_{ij}\left(G  \right) }^{\left(Dh \left( G \right) \right)^{\star} \left[ Z \left( G \right)\right]}, \dot{G}\right\rangle_{\mathcal{F}}
\end{align}

%\subsection{Proof of the corollary to the EMI property}

\subsection{Extra numerical experiments}
We also report the numerical experiments of a simpler toy experiment. In this experiment, the data are generated in the same way :
\begin{align*}
X =& Q\text{diag} (\lambda_1, \ldots, \lambda_r,\mu_1, \ldots, \mu_r ) Q^\top \\
+& V\text{diag} (|\epsilon_1|, \ldots, |\epsilon_{2r}| ) V^\top
\end{align*}
with $Q$ and $V$, two random orthonormal square matrices,
and $\lambda_i \sim \mathcal{N}(5,0.2)$ or $\lambda_i \sim \mathcal{N}(4,0.1)$ respectively for the positive and negative classes 
and $\mu_i \sim \mathcal{U}([1,3])$ and $\epsilon_i \sim \mathcal{N}(0,1)$ independently of the class.

The difference between those two experiments resides in the range in which the variables $\mu_i$ are sampled. Here, the range of the uniform distribution is smaller and this toy dataset becomes easier to classify.

The datasets were composed of $50$ examples in the training set and $500$ in test and every experiment was repeated $10$ times.

\begin{table}
\caption{Second experiment on a toy dataset with a weaker level of noise information.}
\begin{center}
\begin{tabular}{|c||c|c|c||c|c|}
\hline
& \multicolumn{3}{c||}{LogEuclidean} & & \\\cline{2-4} 
size	 & $\mathbb{I}_d$ & $\bar{X}$ &	$G_{\text{KTA}}	$ & 	$\delta_r$& 	$\delta_e$ \\\hline
6x6	&83.22&	83.00&	 89.10 &	83.06&91.56\\\hline
8x8&	85.10&	85.00		& 91.90 &	84.98&94.82\\\hline
16x16&	89.38&	89.54	& 96.88 &	89.46& 98.28\\\hline
20x20&	92.72 &	 92.74	& 98.60 &	 92.74& 99.42\\\hline
\end{tabular}
\end{center}
\end{table}

In this simple experiment, as in the other toy experiment, the Euclidean distance achieves the best performances for the same reasons.

Even if the Euclidean distance seems to be the most suited to those data, this toy experiment is interesting. Indeed, among the distances respecting the Riemannian geometry of the data (namely the LogEuclidean and Riemannian distances), the LogEuclidean distance parametrized by our approach performs the best. Optimizing the reference point $G$ (with respect to a KTA criterion) seems to diminish the impact of the uninformative variables $\mu$.

Note that in this experiment, the results of all approaches (including the Riemannian approaches) improves as the dimensionality grows. Contrary to the results reported in~\ref{tab:toyExpe}, in this experiment, the level of information is higher than the level of noise in the data. Hence, the more dimension we add, the easier it becomes to classify.
\end{document}